\documentclass[conference]{IEEEtran}
\IEEEoverridecommandlockouts
\usepackage{cite}
\usepackage{amsmath,amssymb,amsfonts}
\usepackage{graphicx}
\usepackage{textcomp}
\usepackage{color}
\usepackage{longtable}
\usepackage{algorithm2e}
\usepackage{algpseudocode}
\usepackage{multirow}
\usepackage{array}
\usepackage{amsmath}
\usepackage{graphicx}
\usepackage{adjustbox}
\usepackage{rotating}
\usepackage{tikz}
\usepackage{mathtools}
\usepackage{xcolor}
\usepackage{multirow}
\def\BibTeX{{\rm B\kern-.05em{\sc i\kern-.025em b}\kern-.08em
    T\kern-.1667em\lower.7ex\hbox{E}\kern-.125emX}}

\begin{document}

\title{Predefined domain specific embeddings of food concepts and recipes: A case study on heterogeneous recipe datasets
}

\author{\IEEEauthorblockN{Gordana Ispirova\textsuperscript{1,2}}
\IEEEauthorblockA{\textit{\textsuperscript{1}Computer Systems Department} \\
\textit{Jožef Stefan Institute}\\
Ljubljana, Slovenia \\
\textit{\textsuperscript{2}International Postgraduate School Jožef Stefan}\\
Ljubljana, Slovenia \\
gordana.ispirova@ijs.si}
\and
\IEEEauthorblockN{Tome Eftimov\textsuperscript{1,3}}
\IEEEauthorblockA{\textit{Computer Systems Department} \\
\textit{Jožef Stefan Institute}\\
Ljubljana, Slovenia \\
tome.eftimov@ijs.si}
\and
\IEEEauthorblockN{Barbara Koroušić Seljak\textsuperscript{1,4} }
\IEEEauthorblockA{\textit{\textsuperscript{1}Computer Systems} \\
\textit{Jožef Stefan Institute}\\
Ljubljana, Slovenia \\
barbara.korousic@ijs.si}

}

\maketitle

\begin{abstract}
Although recipe data are very easy to come by nowadays, it is really hard to find a complete recipe dataset -- with list of ingredients, nutrient values per ingredients, and per recipe, allergens, etc. Recipe datasets are usually collected from social media websites where users post and publish recipes. Usually written with little to no structure, using both standardized and non-standardized units of measurement. We collect six different recipe datasets, publicly available, in different formats and some including data in different languages. Bringing all of these datasets to the needed format for applying a machine learning (ML) pipeline for nutrient prediction \cite{ispirova_p-nut_2020,ispirova_domain_2021}, includes data normalization using dictionary based named entity recognition (NER), rule based NER, as well as conversions using external domain specific resources. From the list of ingredients, domain-specific embeddings are created using the same embedding space for all recipes -- one ingredient dataset is generated. The result from this normalization process are two corpora -- one with predefined ingredient embeddings and one with predefined recipe embeddings. On all six recipe dataset the ML pipeline is evaluated. The results from this use case also confirm that the embeddings merged using the domain heuristic yield better results than the baselines. 
\end{abstract}


\section{Introduction}
In this era of social media, internet content containing recipe data are in abundance. However, well-structured datasets containing recipe data, that can be used for research purposes or reused for some application scenario that requires structure, are very rare, almost non-existent. Recipe datasets are most often than not, collected and extracted from food-focused online social networking services, websites and mobile apps that provide recipes to users, which are written from other such users. There are plenty of such services, to name a few -- Allrecipes \cite{all_recipes}, Yummly \cite{yummly}, Epicurious \cite{epicurious}. These recipes are usually written with little to no structure, using both standardized and non-standardized units of measurement.\\
Recipe1M is the only publicly available recipe dataset that has structured data -- separated list of ingredients, quantities and measurements, as well as nutrient values per recipe and per ingredient.
In \cite{ispirova_p-nut_2020} we present a ML pipeline (called P-NUT), for predicting nutrient values of a food item considering learned vector representations of text describing the food item. 
Based on this ML pipeline, in \cite{ispirova_domain_2021} we propose a domain heuristic for merging text embeddings. The evaluation results show that the embeddings merged with the domain heuristic outperformed the embeddings merged with traditional merging techniques as features in the modeling for predicting nutrient values. Recipe1M is the only publicly available recipe dataset that has the necessary data to apply the proposed ML pipeline. The goal of this study is to generate predefined embeddings for recipe data learned using heterogeneous, multi-lingual datasets.\\
Text embeddings have gained a lot of traction and are an essential and standard component of many NLP studies. The principal use has been "transfer learning", where using very big datasets of raw, unlabeled data (from different sources, e.g. web scrapping)  we first learn embeddings, and then use these pre-trained i.e. predefined embeddings as inputs of a model in a supervised task (a classical ML pipeline).\\
In  \cite{howard2018universal} the authors demonstrate the ability to use transfer learning for text data. Transfer learning has been shown to perform exceptionally well for imaging tasks \cite{gulshan2016development,beam2016translating} because of pre-trained computer vision models \cite{simonyan2014very, szegedy2016rethinking, he2016deep}, all of which are pre-trained on the ImageNet database \cite{deng2009imagenet}. \\
In the means of domains, work like this has been done in the medical domain, for example cui2vec \cite{beam2019clinical} -- which are predefined set of embeddings for 108,477 medical concepts. In \cite{choi2016multi} the authors present Med2Vec -- a simple, robust algorithm that learns code and visit representations by using huge EHR datasets with over a million visits. It enables us to interpret the representations that have been positively validated by clinical specialists.\\
In the Food and Nutrition domain, however, there is a lack of this kind of studies.
Existing work that focuses on textual content to learn recipe representations is presented in \cite{li2020reciptor, tian2021recipe}. They take advantage of the instructions and ingredients associated with recipes. Several other studies consider the images \cite{carvalho2018cross, salvador2017learning, wang2019learning, salvador2021revamping} and they typically concentrate on the recipe-image retrieval task and
attempt to align images and text together in a shared space, resulting in information loss for both modalities. In \cite{tian2022recipe2vec}, the authors focus on learning recipe representations using multi-modal information extracted from images, text, and relations, focusing on cuisine category classification and region prediction, based on the recipe categories, meaning the textual data included is only through the recipe category.\\
The task of predicting nutrient values in this context still remains unexplored. Moreover, recipe embedding based on the quantities of the ingredients have never been introduced prior to our work in \cite{ispirova_domain_2021}. In this study, we collect recipe data from six heterogeneous recipe datasets: Indian recipes, Recipe1M, Epicurious, Salad recipes, Yumlly28K, Recipe box. The datasets are first brought to the same format, through the processes of named entity recognition (NER) and data mapping. We use dictionary-based NER for extracting the measurement units and rule-based NER for extracting the quantities of the ingredients. The data mapping process is an ensemble of multiple approaches, which we use for mapping the ingredients to a Food Composition Database (FCDB) from the United States Department of Agriculture (USDA) \cite{department_of_agriculture_usda_2021} called FoodData Central USDA to obtain the nutrient values. \\
The end product from this study are two predefined embedding corpora -- for ingredient vector representations and recipe vector representations, as well as four domain dictionaries constructed with and approved by a domain expert: a dictionary for units of measurement, for converting units of measurement, for names of branded foods, and a dictionary for redundant words specific to recipe data.
Training embeddings tailored for a specific task is a very time consuming process, therefore the corpora of predefined embeddings can be used for research purposes as well as for application purposes transferring them to other tasks.\\
The rest of the paper is structured as follows: in Section \ref{Methods} first, we give the related work and a detailed explanation of the data for the experiments and the methodology. The experimental results, evaluations and further discussions are presented in Section \ref{results}. At the end, in Section \ref{Conclusion} a summarization of the importance of this methodology and directions for future work are presented.

\section{Methods}
\label{Methods}
This section begins with a review of the state-of-the-art needed to understand the methodology, the algorithms used, and recent work done in this area; the section continues with a description of  data used in the experiments and it ends with a detailed explanation of the methodology.

\subsection{Related work}
\subsubsection{FoodEx2 Classification System}
FoodEx2 \cite{european_food_safety_authority_food_nodate} is a thorough method for classifying and describing foods that aims to address the requirement for food descriptions in data collections across several food safety domains. 
FoodEx2, a standardized method for categorizing and characterizing foods, is a supplement to the Standard Sample Description (SSD2) \cite{european2013standard} (data model to describe food and feed samples and analytical results), and it is made up of descriptions of numerous different food items that have been combined into food groups and larger food categories in a parent-child hierarchy.

\subsubsection{StandFood}
\label{standfood}
StandFood \cite{eftimov_standfood_2017} is a method that standardizes foods according to FoodEx2 classification system, and it consists of three parts.
The first part uses a machine learning approach to classify foods into four FoodEx2 categories: raw (r) and derivatives (d) for single foods, and simple (s) and aggregated (a) for composite foods (c). To describe foods, the second uses a natural language processing approach combined with probability theory. In order to improve the classification result, the third component of the StandFood method integrates the results from the first and second parts by creating post-processing criteria. 
The StandFood evaluation results demonstrate that the system produces promising results and may be used to classify and describe food items in accordance with FoodEx2. FoodEx2 codes can be found missing in food composition databases and food consumption data using StandFood, allowing users to compare and combine them.

\subsubsection{USDA Food Composition Database}
\label{USDAsubsection}
FoodData Central is the The Food Composition Database (FCDB) from the United States Department of Agriculture (USDA) and it is a comprehensive, research-focused data system that offers links to sources of information about agriculture, food, dietary supplements, and other topics in addition to increased data on nutrients and other food components. 
The need for open and easily accessible information about the nutrients and other ingredients in foods and food products has significantly increased in recent years due to the rapidly expanding pace of change in the food supply and the growing diversity of uses for food data. In order to analyze, gather, and present dietary profile data in a way that is rigorously scientific, a new methodology is needed. FoodData Central is still the embodiment of this novel strategy.
From the FoodData Central USDA FDCB four datasets are of our importance: 

\begin{itemize}
    \item [a.] Foundation Foods -- contains values determined from assessments of nutrients on individual samples of commodity/commodity-derived minimally processed foods with insights into variability, as well as significant underlying metadata. 
    \item [b.] Nutrient Database for Dietary Studies (FNDDS) -- Data on nutrients and portion weights for foods and beverages reported in What We Eat in America, the dietary intake section of the National Health and Nutrition Examination Survey (NHANES) \cite{NHANES}. 
    \item [c.] Branded Foods -- Data from labels of national and international branded foods collected by a public-private partnership. 
    \item [d.] Standard Reference (SR) Legacy -- Historic data on food components including nutrients derived from analyses, calculations, and published literature. 
\end{itemize}

\subsubsection{Domain dictionaries}
\label{dictionaries}
In order to perform extraction of the needed information from the text in the recipe datasets, there is a need of domain dictionaries -- specifically, a dictionary with units of measurements (from the International System of Units (SI) and household measurements). Unfortunately, there were no available dictionaries that would be useful for our case, therefore we proceeded with assembling the needed resources with the assistance of a domain expert -- a nutritionist.

\begin{itemize}
    \item [a.] Units of Measurement Dictionary: This dictionary is constructed with the help of a domain expert -- a nutritionist, and it includes possible units of measurement for different food items that can be found in recipe data:
\begin{itemize}
    \item Units of measurement for quantity from the International System of Units (SI) -- these measurements are official and defined in \cite{gHobel2006international, newell2019international}, and are established and maintained by the General Conference on Weights and Measures (CGPM) \cite{turner73interpretation}. Example: grams, milligrams, etc.
    \item Units of measurement for quantity from the Household Measurement System -- which is not an official system for measuring units, but it is most commonly used in everyday life. Example: tablespoon, teaspoon, cup, etc.
    \item Units of measurement for quantity from the Apothecaries' system -- which originated as the system of weights and measures for dispensing and prescribing medications, is a historical system of mass and volume units that were used by physicians and apothecaries for medical prescriptions and also sometimes by scientists \cite{kruger1830complete, henry1831elements, gehler1844johann}. Example: pint, dram, scruple, etc.
    
\end{itemize}

The dictionary of Units of Measurement contains 144 instances. 

\item [b.] Redundant Words Dictionary:
This dictionary includes a list of words and phrases that can be considered as redundant in this scenario. These can be -- words that do not bring any weight or value to the nutritional content of the ingredient in question, as well as additional explanation that the person that wrote the recipe included in order to further explain either the ingredient itself or the cooking process. This dictionary contains 415 words and phrases in total. Examples with redundant words/phrases: "1 lb large shallots, \textit{bulbs separated if necessary and each bulb halved lengthwise}",
"12 oz \textit{best available quality} salmon\textit{, cut into paper thin pieces and arranged} \textit{in rows in a container, separated, and left in the refrigerator for half an hour}".
\item [c.] Branded foods dictionary:
This dictionary includes popular food brand names that can often replace the word for the actual food item they produce. This dictionary is extracted from the branded food dataset from the USDA FCDB. Example of such food items are: "M\&Ms", "Cola", "Stevia", "Oreo", "Nescafe", "Dr. Peppers", "7Up", etc.

\item [d.] Conversion Dictionary.\\
This dictionary is constructed using the above mentioned systems of measurements, multiple online conversion calculators, and the opinion and suggestions of an expert. The purpose is to convert all the quantities in one unit of measurement -- grams. This dictionary contains conversions for all instances contained in the Units of Measurement dictionary.

\end{itemize}
\subsubsection{P-NUT Predicting Nutrient Values using Text Descriptions}
P-NUT \cite{ispirova_p-nut_2020} is a novel ML pipeline for learning predictive models that incorporates the domain knowledge encoded in external semantic resources. 
The ML pipeline (P-NUT) consists of three parts:
\begin{enumerate}
    \item [a.] Representation learning: Introduced by Mikolov et al. in 2013 \cite{mikolov_distributed_2013} and Pennington et al. in 2014 \cite{pennington_glove_2014}, word embeddings have become indispensable for natural language processing (NLP) tasks in the past couple of years, and they have enabled various ML models that rely on vector representation as an input to benefit from these high-quality representations of text input. This kind of representation preserves more semantic and syntactic information of words, which leads to their status as being state-of-the-art in NLP.
    \item [b.] Unsupervised machine learning: Nutrient content exhibits notable variations between different types of foods. In a big dataset, including raw/simple and composite/recipe foods from various food groups, the content of nutrients can have values from 0–100 g per 100 g.
    \item [c.] Supervised machine learning part: The final part of the P-NUT methodology is supervised ML, where separate predictive models are trained for the nutrients that we want to predict. The nutrient values are continuous data; therefore, the models are trained with single-target regression algorithms, in which, as input, we have the learned embeddings of the short text descriptions, clustered based on the chosen/available domain-specific criteria. Selecting the right ML algorithm for the purpose is challenging; the default accepted approach is selecting a few ML algorithms, setting the ranges for the hyper-parameters, hyper-parameter tuning, utilizing cross-fold validation to evaluate the estimators' performances (with the same data in each iteration), and at the end, benchmarking the algorithms and selecting the best one(s). The most commonly used baselines for regression algorithms are the central tendency measures, i.e., mean and median of the train part of the dataset for all the predictions.
\end{enumerate}

\subsubsection{Domain Heuristic for Merging Textual Embeddings}
Using the concept of our proposed ML pipeline presented in \cite{ispirova_p-nut_2020} and \cite{ispirova_exploring_2020} we constructed a representation learning pipeline (presented in \cite{ispirova_domain_2021}) in order to explore how the prediction results change when, instead of using the vector representations of the recipe description, we use the embeddings of the list of ingredients. The nutrient content of one food depends on its ingredients; therefore, the text of the ingredients contains more relevant information. We define a domain-specific heuristic for merging the embeddings of the ingredients, which combines the quantities of each ingredient in order to use them as features in machine learning models for nutrient prediction. The results from the experiments indicate that the prediction results improve when using the domain-specific heuristic. The prediction models for protein prediction were highly effective, with accuracies up to 97.98\%. Implementing a domain-specific heuristic for combining multi-word embeddings yields better results than using conventional merging heuristics, with up to 60\% more accuracy in some cases. 

\subsection{Data}
\label{datasets}
For this study, all publicly available recipe datasets were explored, and six others, besides Recipe1M, that had the potential to be transformed into the necessary format were selected. 
The selected datasets including Recipe1M are:\\
\begin{enumerate}
    \item [a.] Recipe1M -- contains 51,500 recipes and the following data for each: recipe title (short textual description of the recipe), structured list of ingredients, recipe instruction, nutrient content of ingredients (quantity in grams of fat, protein, saturates, sodium, and sugar per 100 grams of the ingredient for each ingredient), quantity of each ingredient, units of measurement per each ingredient (household measurement system), weight in grams per each ingredient, nutrient content (quantity in grams of fat, protein, salt, saturates, and sugars per 100 grams of the recipe), FSA traffic light labels per 100 grams.
    \item [b.] Indian recipes -- contains 6,871 recipes and the following data for each: recipe url, raw continuous text with ingredients, quantities and units of measurement in Hindi, recipe instruction.
    \item [c.] Epicurious -- contains 20,103 recipes and the following details for each: recipe title, recipe url, raw continuous text with ingredients, quantities and units of measurement, calorie content, nutritional values for protein, fat and sodium.
    \item [d.] Salad recipes -- contains 82,243 recipes and the following details for each: recipe title, recipe instruction, raw continuous text with ingredients, quantities and units of measurement.
    \item [e.] Yummly28k -- contains 27,639 recipes and the following data for each: recipe title, raw continuous text with ingredients, detailed nutrient information – nutrient values for 94 nutrients.
    \item [f.] RecipeBox -- contains 39,802 recipes and the following data for each: recipe title, raw continuous text with ingredients, quantities and units of measurement, recipe instruction.
\end{enumerate}

\subsection{Methodology}
\subsubsection{Data normalization}
After collecting the datasets, they need to be transformed into the needed format in order to apply the ML pipeline. This process is separated into two parts: extracting the needed information -- with dictionary-based NER extracting the unit/s of measurement and with rule-based NER extracting the quantity of the ingredient; and data mapping -- for mapping the ingredient (food item) to the USDA FCDB.

\begin{itemize}
\item [a.] Extracting Information from Unstructured Recipe Data -- Before any step is taken towards transforming the datasets into the needed format, it is important to note that two of the datasets contain data that is not in English -- Indian recipes and Yummly28K. This means that we are dealing with multilingual data. In order to obtain the translation of the data the Google Translate API in Python \cite{han2020googletrans} is used. 
The results obtained in this way are reliable because we are dealing with one sentence text, that carries little to no context. \\
Now, let us define the ideal format that is needed for applying the ML pipeline.
Preferably, we would have dataset $D$ with $n$ number of instances/recipes and $m$ number of features representing those instances. In those $m$ features out of our importance is that we have the following:
\begin{itemize}
    \item Name of recipe;
    \item Nutritional information about the whole recipe (on 100 grams);
    \item List with names of ingredients;
    \item List with quantities in grams for each ingredient (ideally on 100 grams);
\end{itemize}
Unfortunately, none of the datasets come in this format, only Recipe1M has the name of the ingredients and the quantities separately. The other five dataset all have raw continuous text with ingredients, quantities and units.
Therefore, the first step is for each recipe in each dataset to separate the names of the ingredients, the quantities and the units into three separate lists. This process would be just a simple task of text splitting i.e. string splitting, which can be performed using a simple function/s in Python \cite{python3stringsplit} and search patterns using regular expression \cite{aho1990algorithms,mitkov2022oxford}, if this text is following some kind of a structure. An example of how this would work is given in Table \ref{separated}).

\begin{table}
\begin{center}
\caption{Example of well structured not separated ingredient list in a recipe dataset.}
\resizebox{\columnwidth}{!}{
\centering
\begin{tabular}{|c|c|}
\hline
\textbf{Example} & \parbox{5cm}{\centering 2 cups flour, whole wheat; \\ 15 grams baking powder; \\ 300 milliliters 3.5\% milk; \\ 2 tablespoons cocoa powder.}\\ \hline
\textbf{List of quantities} & $\textbf{[}2,15,300,2\textbf{]}$ \\ \hline
\textbf{List of units} & $\textbf{[}$cups, grams, milliliters, tablespoons$\textbf{]}$ \\ \hline
\textbf{List of names of ingredients} & \parbox{5cm}{\centering $\textbf{[}$flour whole wheat, \\ baking powder, \\ 3.5\% milk, \\ cocoa powder$\textbf{]}$} \\
\hline
\end{tabular}}
\label{separated}
\end{center}
\end{table}

When inspecting the datasets, more specifically the raw continuous text that contains the list of ingredients, quantities and units, it is more than evident that this technique will not be enough in this case, in view of the fact that the text is not following any pattern or rule. The text of our interest in all of the recipe datasets did not have any kind of structure. There were a few things that should be taken into account:
\begin{itemize}
    \item The measurement units:
    \begin{itemize}
    \item Can be of different type:
    \begin{itemize}
        \item from the The International System of Units (SI) i.e. the metric system (e.g. grams, milliliters);
        \item a household measurement (e.g. cup, tbs);
        \item an approximate measurement (e.g. splash, pinch, dash, handful, thumb sized, smidgen);
        \item specific for a single food or a group of foods (e.g. "clove of garlic", "head of lettuce", "a cinnamon stick", "a stick of butter").
    \end{itemize}
    \item Can be written in different ways:
    \begin{itemize}
        \item with full text (e.g. grams, tablespoon);
        \item different type of abbreviations (e.g. for tablespoon -- tb, tbs, tbsp; for pound -- lb, lbs, lbss);
    \end{itemize}
    \item Can be non existent (e.g. "3 eggs", "2 onions")
\end{itemize}
\item The quantity:
\begin{itemize}
    \item Can be written in many different ways:
    \begin{itemize}
        \item with numbers (e.g. "250 ml orange juice");
        \item with words (e.g. "two apples");
        \item a combination of number and words (e.g. "4 and a half tbsp sugar", "Two 0.16 ounces packets instant noodles" );
        \item as indefinite articles (e.g. "a cup of milk");
    \end{itemize}
    \item Can be non existent (e.g. "salt and pepper").
    \item Can be "user-defined" (e.g. "cinnamon, to taste").
\end{itemize}
\item The text representing the ingredient:
\begin{itemize}
    \item Can be a simple food (e.g. "2 pounds cherries");
    \item Can be a branded food (e.g. "1 (40 grams) packet M\&M's chocolate candies);
    \item Can have additional explanation about the food itself (e.g. "1 lb large shallots (8), bulbs separated if necessary and each bulb halved lengthwise", "½ pound skinless chicken breast (about 1 chicken breast, cut into ½-inch dices)");
    \item Can be something that is not food but it is written in the ingredient list as it is needed for the recipe (e.g. "1 sheet of parchment paper").
\end{itemize}
\end{itemize}

The first attempt is to formulate all these cases and put together rules for string splitting, including rules for extracting the quantity and extracting the unit of measurement. 
\begin{itemize}
    \item Extracting the measurement unit/s -- In order to extract the unit of measurement from the text, a dictionary was constructed, defined in subsection \ref{dictionaries} -- therefore, every entity or entities from that dictionary found in the text of matter is defined and extracted as a unit of measurement. 
    \item Extracting the quantity/quantities -- In order to extract the quantity from the text, we define certain rules and patterns:
    \begin{itemize}
        \item If there are digit occurrences in the format     $D_{integer} \frac{D_{integer}}{D_{integer}}$. Where $D$ can be one or multi -- digit integer number. This is summed into one number.
        \item If there are digit occurrences in the formats: $D_{integer}\;D_{integer}$, $D_{integer}\;D_{float}$, $D_{float}\;D_{integer}$, $\frac{D_{integer}}{D_{integer}}\;\frac{D_{integer}}{D_{integer}}$, $\frac{D_{integer}}{D_{integer}}\;D_{integer}$, $D_{float}\;D_{float}$. Where $D_{integer}$ can be one or multi -- digit integer number, and $D_{float}$ is a two or multi digit float number. If there is an occurrence like this, the numbers from the occurrence are multiplied.
    \end{itemize}
    \item Extracting the food item -- In order to extract the food item on the text that remains after the extraction of the measurement unit/s and quantity/quantities we use the \textit{Redundant words dictionary}, and remove any occurrences of any of the words belonging in the dictionary. Everything that is left is the food item. 
\end{itemize}

There are a few special cases where some words can be considered as measurement units in some cases and be omitted in others. When either one of the following: "jar", "can", "packet", "package", "box", "bottle", and "container", appears in the text representing an ingredient there are two scenarios:
\begin{itemize}
    \item If is proceeded with a measurement unit from the defined \textit{Units of Measurement dictionary}, it is omitted (e.g. "1 (500 grams) container Greek yogurt").
    \item If it is found in a text where no measurement unit from the defined dictionary is extracted, it is considered a unit of measurement (e.g. "2 packets of Stevia").
\end{itemize}
For some food items there are specific units of measurement, that correspond only to them. These words are extracted as units of measurement only when they appear in pair with the food item/s they are correlated with. These word pair are:
\begin{itemize}
    \item "clove" and "garlic";
    \item "stick" and "butter"/"margarine"/"cinnamon"/
    "carrot"/"celery";
    \item "sprig" and "rosemary"/"thyme"/"mint"/"parsley";
    \item "link" and "sausage";
    \item "stalk" and "celery"/"green onion"/"spring onion"/
    "broccoli"/"kale"/"cauliflower";
    \item "sheet" and "gelatin";
    \item "cube" and "stock"/"butter","margarine";
    \item "head" and "cabbage"/"lettuce"/"cauliflower";
\end{itemize}

Each of these, of course, have different weights for the different food items, this is taken into account in the \textit{Conversion dictionary}.\\
Therefore, we can see that the process of extracting the unit/s of measurement is a hybrid process of NER with dictionary and additional rules.
These rules, when applied, should result in separated lists of units of measurement (one or more), quantities (one or more) and food item. However, after defining these steps we stumbled upon more difficulties with the process of extracting the needed information:, to be more specific -- text designated for one ingredient in the list of ingredients can contain more than one ingredient (concatenated with the conjunction "plus", "and" or "with"), contain multiple options for one ingredient (concatenated with the conjunction "or"), contain more than one quantity for one ingredient (concatenated with the preposition "to"), or a combination of two or more of the aforementioned. 
The solutions that were deemed fit for these cases are:
\begin{itemize}
    \item Containing conjunction "and" or "with":
    \begin{itemize}
        \item If the condition: $(count("and") \geq 1 \lor count("with") \geq 1) \land count("plus") == 0$ is satisfied and there are two units of measurement extracted on each side of the conjunction, the conjunction "and" is replaced with "plus" and is treated as such forward (e.g. "2 4 oz packages of salmon and 5 tablespoons of lemon juice").
        \item Any other cases with the conjunction "and" are dismissed (e.g. "1 lb large shallots (8), bulbs separated if necessary and each bulb halved lengthwise").
    \end{itemize}    
    \item Containing conjunction "plus":
    \begin{itemize}
        \item If $count("plus") \geq 1$ -- split on "plus" and separate text to as many ingredients as the count of "plus" in the text (e.g. "1/4 cup finely chopped sweet gherkins plus 2 tablespoons pickled juice from the jar, plus 12 whole gherkins").
    \end{itemize}
    \item Containing conjunction "or":
    \begin{itemize}
        \item If $count("or")>1$ -- split on "or" and separate text to two part, but keep the first part only;
        \item If $count("or")>1$ -- check the position of the extracted unit/s of measurement, split into as many pars as the number of "or" and:
        \begin{itemize}
            \item If there is one unit of measurement extracted, take the part with the unit of measurement.
            \item If there is more than one unit of measurement, all or none of them with assigned quantity, and there is no "plus" in the ingredient take the part with the first unit of measurement.
            \item If there is more than one unit of measurement, some with unassigned quantity, and there is no "plus" in the ingredient take the part with unit of measurement with assigned quantity.
        \end{itemize} 
    \end{itemize}
    \item Containing preposition "to":
      \begin{itemize}
        \item If "to" is in a specific pattern involving integer, and/or float numbers, and/or digits separated with "/" (e.g.: "2 14 1/2- to 15-ounce cans diced tomatoes in juice"), the first part of the pattern when split by the word "to" is kept and appended to the rest of the text.
    \end{itemize}  
\end{itemize}

Despite, these cases we determined that there are other cases with occurrences of two ingredients expressed within text designated for one, to be more specific, text containing any of the following phrases: "mixed with", "mixed in", "beaten with", "dissolved in", "sauteed in", "diluted", "combined", or the verb "add". This is resolved by treating and replacing these phrases with "plus".
Some examples are: "5 to 6 anchovies \textbf{add} 1 teaspoon red pepper flakes", "1 large egg \textbf{beaten with} pinch of salt", "1 teaspoon corn starch \textbf{dissolved in} 2 tablespoons milk".

Another problem that is reoccurring is the appearance of more than one quantity for one ingredient, meaning having information that is extra in the text. This is occurring in multiple scenarios and for each of these scenarios a fitting solution is applied:
\begin{itemize}
    \item If we extract two different patterns for quantity and one unit of measurement after the second pattern and there is no text between the two patterns that express quantity -- The second pattern is kept as the quantity, and the first is removed from the text.
    \item If we extract two different patterns that express quantity and one unit of measurement after the first pattern, immediately followed by the second pattern -- The first pattern is kept as the quantity and the second pattern is removed.
    \item If we extract two different patterns that express quantity, and two different units of measurement and there is no text between the two patterns except the unit of measurement connected with the first pattern -- The first pattern is kept as the quantity of the ingredient.
    \item If we extract two different patterns that express quantity, and there is no unit of measurement extracted and there is no text between the two patterns -- The first pattern is kept as the quantity of the ingredient.
    
\end{itemize}

Before applying any rules, the text is pre-processed. For the pre-processing a few rules are followed:
\begin{itemize}
    \item The text is tokenized and lemmatized.
    \item Brackets and redundant words from the \textit{Redundant words dictionary} are removed.
    \item Words from the \textit{Branded food dictionary} are identified, and no further pre-processing is done on them.
    \item Special characters are removed except:
    \begin{itemize}
        \item "\%" -- if it is preceded with a number.
        \item "\&" -- it is replaced with "and".
        \item "+" -- it is replaced with "plus".
        \item "$/$" -- if it is in between digits.
        \item "-" -- is replaced with "to" if it is between digits.
    \end{itemize}
    \item Words depicting numbers are changed with the number/s they represent (special cases are added for "half a", "half of", "half from" replaced with "1/2").
    \item If the text starts with "a/an" followed by a unit of measurement, the "a/an" is replaced with "1".
    \item If the text starts with "a/an" and there is no unit of measurement present and there is no pattern depicting quantity present in the text, the "a/an" is replaced with "1".
    \item If there is "." or "," present after the extracted quantity and measurement and there is no "and/or" in the text, everything after the dot or comma is removed.
\end{itemize}

After all of this is applied to all the instances from the datasets we have the new six new datasets in the format needed for generating the embeddings -- the Representation Learning part of the methodology presented in \cite{ispirova_domain_2021}. 

\item [b.] Data mapping to USDA FCDB --
In order to apply the supervised ML part, we first need to determine which nutrient values we want to predict. We can see from the descriptions of the datasets, given in Subsection \ref{datasets}, that only three datasets contain nutrient values, and all the same nutrient values. For the purpose of obtaining the nutrient values for all datasets, we went the route of mapping the data to a FCDB, and the FCDB of choice is the USDA FCDB \cite{department_of_agriculture_usda_2021}, which made sense not only because the datasets come mostly from websites that are based in the USA, but also for the fact that it is the biggest integrated data system that provides expanded nutrient profile data and links to related agricultural and experimental research. 
This database is used for two goals:
\begin{enumerate}
    \item Obtaining the quantities in grams that are missing in the datasets -- These quantities are missing because there was no unit of measurement present, just the food item and/or a number (e.g.: "2 eggs", "3 apples", "1 standard size chicken breast", etc.).
    \item Obtaining nutrient values for each ingredient, in order to calculate the nutrient values of the recipe.
\end{enumerate}

Before the data mapping process, a new dataset is constructed, consisted of all the unique ingredients extracted from all the six recipe datasets. In total this dataset contains 71,641 instances.
The two aforementioned goals where both achieved with mapping the ingredients to a food item in the USDA database. We have dealt with this type of mapping in previous studies \cite{ispirova_exploring_2020, ispirova2022cafeteriafcd} where we used a lexical similarity approach involving Part of Speech (POS) tagging and probability theory.
This method is limited because of its nature to base the matching on intersection of nouns, which the POS tagging method may not capture. More over, the instance may not contain any nouns or the POS tagging method may not tag the tokens (words) correctly (i.e. the nouns may be tagged as other POS tags, for example "orange" may be tagged as a adjective instead of a noun). In this study, it was of real importance, to obtain matches for as many as possible of the ingredients, therefore we could not rely only on this method. We do incorporate this method in the mapping -- we include the nouns, verbs, adjectives, and numbers and reformulate the similarity as shown in Equations (\ref{e1}), (\ref{e2}), and (\ref{e3}).
\begin{equation}
\label{e1}
p_{n}=\frac{Nouns_i \cap Nouns_j}{Nouns_i \cup Nouns_j}
\end{equation}

\begin{align}
  \label{e2}
  \begin{split}
p_{v+a+num}=
\big( ((Verbs_i\cup Adjectives_j \cup Numbers_i) \cap \\ (Verbs_j \cup Adjectives_j \cup Numbers_j))+1 \big) \\ /\ \big( ((Verbs_i\cup Adjectives_j \cup Numbers_i) \cup \\ (Verbs_j \cup Adjectives_j \cup Numbers_j))+2 \big)
\end{split}
\end{align}

\begin{equation}
\label{e3}
similarity_{index}=p_{n} \times p_{v+a+num}
\end{equation}
Where the set of nouns, verbs, adjectives and numbers from the ingredient we want to match are: $Nouns_i$, $Verbs_i$, $Adjective_i$, $Numbers_i$ and from food item from USDA: $Nouns_j$, $Verbs_j$, $Adjective_j$, $Numbers_j$.

For performing the mapping from the USDA FCDB we extracted the food items from all four datasets from USDA. The initial merged dataset contained 298,318 items, after removing duplicates, the final merged dataset contains 290,519 food items. This dataset is then submitted to some pre-processing: the text is tokenized and lemmatized, brackets and redundant words from the \textit{Redundant words dictionary} are removed.

The two datasets -- with the unique ingredients from the six recipe datasets and the pre-processed food items from the four datasets from the USDA FCDB are then inputs to the data mapping procedure. 
The mapping procedure, is an ensemble of of multiple similarity measures sorted in order of importance. Two food items -- one from USDA and one ingredient from the datasets are considered a match one of the following similarity measures is satisfied (in this particular order of importance):
\begin{itemize}
    \item Same set of lemmas.
    \item Same set of nouns and $similarity_{index}>0$.
    \item The intersection of nouns is not an empty set and $similarity_{index}>0$.
    \item The intersection of the unions of nouns, verbs, adjectives and numbers of the two food items is not an empty set and $similarity_{index}>0$.
    \item The Levenshtein distance \cite{levenshtein1966binary} of the concatenated lemmas of both food items is smaller than the length of the concatenated lemmas of the ingredient that is being matched.
\end{itemize}

\end{itemize}

\section{Results and discussion}
\label{results}

This section contains the evaluation of the presented methodology -- the experimental setup, the results obtained, as well as a discussion about the outcome of the experiments.

\subsection{Predefined Corpus of Ingredient Embeddings}
\label{ingredientembeddingscorpus}
The results from the data normalization process are six harmonized recipe datasets with a structured format, mapped to a FCDB.
After the procedure of extracting information from the unstructured recipe datasets, a dataset with 71,641 ingredients is constructed. On this dataset ingredient embeddings are generated using four embedding algorithms and the following parameters:
\begin{enumerate}
        \item [a.] Word2Vec -- architectures: CBOW and SG, vector dimension: $50$, $100$, and $200$, sliding window: $2$,$3$, $5$, and $10$, and heuristics for combining the individual word embeddings: sum and average.
        \item [b.] GloVe -- vector dimension: $50$, $100$, and $200$, sliding window: $2$,$3$, $5$, and $10$, and heuristics for combining the individual word embeddings: sum and average.
        \item [c.] Doc2Vec -- architectures: PV-DM and PV-DBOW, vector dimension: $50$, $100$, and $200$, sliding window: $2$,$3$, $5$, and $10$, and heuristics for combining the individual word embeddings: sum and average.
        \item [d.] BERT -- combining the last four layers of the neural network: summing and concatenating. 
\end{enumerate}

The result is a predefined corpus of ingredient embeddings.

\subsection{Predefined Corpus of Recipe Embeddings}
After the mapping of the dataset with ingredients to the USDA FCDB, the information about nutrient values form the USDA FCDB is extracted and added alongside each ingredient, that is -- nutrient values for $100$ grams of the ingredient. The missing quantities are extracted from the portions sizes (expressed in grams) of the food items from the USDA FCDB. Therefore, every needed data are obtained in order to calculate the recipe embeddings with the domain heuristic. \\
To generate the recipe embeddings, first we obtain the nutrient values of the recipe per $100$ grams, by calculating the quantity of each ingredient per $100$ grams of the recipe and scaling the nutrient values accordingly. For our target values we selected the following five nutrients: fat, protein, sugar, saturated fat, and sodium. This choice is obvious since the three datasets that have nutrient values, have all or some of these. \\
Using the predefined corpus of ingredient embeddings and the calculated nutrient values with the mapping to the USDA FCDB we generate the recipe embeddings for the six recipe datasets by combining the ingredient embeddings with the domain heuristic presented in \cite{ispirova_domain_2021}. This way of generating the domain specific recipe embeddings -- with the same dataset of ingredient embeddings, brings them all in the same landscape i.e. the same feature space.
The end result are six datasets of recipe embeddings, which are combined to form the predefined corpus of recipe embeddings, consisting of 219,765 recipe embeddings.

\subsection{Predictive modeling}
After obtaining the embeddings, the next step is the predictive modeling, single-target regressions with six different types of regressions (Linear, Ridge, Lasso, Elastic Net, Decision Tree, Random Forest, and Neural Network regression) for predicting five nutrients (fat, protein, sugar, saturated fat, and sodium) using four different types of embedding algorithms (Word2Vec, GloVe, Doc2Vec, and BERT). In total there were 3,660 models trained: 1,440 Word2Vec, 720 GloVe, 1,440 Doc2Vec, and 60 BERT models. 
The predictive models are evaluated with the next steps:
\begin{enumerate}
    \item  [a.] Hyper-parameter tuning -- from the scikit-learn library in Python \cite{pedregosa_scikit-learn_2011}): GridSearchCV (all parameter combinations) for Linear, Ridge, Lasso, Elastic Net, and Decision Tree regression, and RandomizedSearchCV (sample a given number of candidates from a parameter space) for Random Forest and Neural Network, because they are more complex algorithms and the exhaustive search from GridSearchCV requires considerably much longer execution time.
    \item [b.] Training with the best parameters from the hyper-parameter tuning.
    \item [c.] K-fold cross-validation to estimate the prediction error.
    \item [d.] Calculating the domain specific accuracy defined in \cite{ispirova_p-nut_2020,ispirova_exploring_2020,ispirova_domain_2021} and the baselines -- mean and median.
\end{enumerate}

For comparing the results, we also trained models with the recipe embeddings obtained without the domain heuristic for merging. We present the results obtained with the domain heuristic and without it for each of the six datasets: Recipe1M, Indian recipes, Epicurious, Salad recipes, Yummly28K, and Recipe box in tables \ref{Rrecipe1mDH}, \ref{RepicuriousDH}, \ref{RsaladrecipesDH}, \ref{Ryummly28KDH}, \ref{RsaladrecipesDH}, and \ref{RrecipeboxDH}.


\section{Conclusion}
\label{Conclusion}
With this study we achieved harmonization over the meta-data of six different heterogeneous recipe datasets, as well as provide the research community with two corpora of predefined domain specific embeddings (one with ingredient embeddings and one with recipe embeddings), and four different domain specific dictionaries, created with the help of a domain expert -- a nutritionist: dictionary for units of measurement, dictionary for converting units of measurement to grams, dictionary for branded foods, and dictionary for redundant words specific to text from recipe data.\\
Predefined domain specific embeddings, as discussed previously, already exist in several domains, as well as the Food and Nutrition domain, but none for recipe data that includes external domain resources, and a domain heuristic for merging. The predefined corpus of ingredient embeddings is currently, the biggest existing corpus of food items that are part of recipes -- as ingredients. This corpus, as a standalone predefined corpus of embeddings can easily be used and transferred into other ML tasks and studies beside the task presented here -- predicting nutrient values. The predefined corpus of recipe embeddings is the largest corpus of recipe embeddings, consisting of domain specific embeddings for 228,158 different recipes, which makes it a powerful resource in the Food and Nutrition domain, and the ML community, as training new embeddings for a specific ML task is a demanding and timely process. These embeddings, as the ingredient corpus can be transferred to other ML tasks and used in other research studies.\\
We also evaluated the effectiveness of the ML pipeline and the domain heuristic for merging multi-word embeddings, the results showed that the ML pipeline is not dataset biased and can perform on heterogeneous data. The diversity of the data in these datasets also provides a base for generalizing the prediction models to many different scenarios with minimum amount of data.\\

\begin{table}
\centering
\caption{Results for the Recipe1M dataset.}
\label{Rrecipe1mDH}
\resizebox{\columnwidth}{!}{
\begin{tabular}{|c|c|c|c|c|c|}
\hline
\textbf{Embedding} & \textbf{Target} & \textbf{\begin{tabular}[c]{@{}c@{}}Average\\ Accuracy\\ (DH)\end{tabular}} &
  \textbf{\begin{tabular}[c]{@{}c@{}}Mean\\ Average\\ Accuracy\end{tabular}} &
  \textbf{\begin{tabular}[c]{@{}c@{}}Median\\ Average\\ Accuracy\end{tabular}} &
  \textbf{\begin{tabular}[c]{@{}c@{}}Average\\ Accuracy\\ (no DH)\end{tabular}}\\
\hline \multirow{5}{*}{Word2Vec} & Fat & 67.86 & 7.86 & 7.48 & 21.88\\
                          & Protein & 86.22 & 27.14 & 22.89 & 62.17 \\
                          & Saturated fat & 85.35 & 7.27  & 8.78 & 49.69\\
                          & Sugars& 60.76 & 6.96  & 13.47 & 25.73\\
                          & Sodium & 91.00 & 25.42 & 43.58 & 81.14 \\
\hline \multirow{5}{*}{GloVe}    & Fat & 62.96 & 7.84 & 7.53 & 21.46\\
                          & Protein & 83.95 & 27.06 & 22.89 & 60.91\\
                          & Saturated fat & 81.69 & 7.27  & 8.78 & 49.54 \\
                          & Sugars & 61.5  & 6.94  & 13.47 & 26.13 \\
                          & Sodium & 90.5  & 25.39 & 43.57 & 81.06\\
\hline \multirow{5}{*}{Doc2Vec}  & Fat & 62.53 & 7.84 & 7.53 & 20.57\\
                          & Protein & 82.6  & 27.06 & 22.89 & 59.76\\
                          & Saturated fat  & 81.9 & 7.27  & 8.78 & 48.05\\
                          & Sugars & 55.62 & 6.94 & 13.47 & 22.47\\
                          & Sodium & 89.6  & 25.39 & 43.57 & 80.93\\
\hline \multirow{5}{*}{BERT}     & Fat & 77.64 & 7.84  & 7.53  & 26.29\\
                          & Protein & 89.32 & 27.06 & 22.89 & 63.57\\
                          & Saturated fat & 90.95 & 7.27 & 8.78 & 55.08 \\
                          & Sugars & 85.64 & 6.94 & 13.47 & 33.44 \\
                          & Sodium & 92.06 & 25.39 & 43.57 & 81.19 \\
\hline
\end{tabular}%
}
\end{table}

\begin{table}
\centering
\caption{Results for the Indian recipes dataset.}
\label{RindianrecipesDH}
\resizebox{\columnwidth}{!}{
\begin{tabular}{|c|c|c|c|c|c|}
\hline
\textbf{Embedding} & \textbf{Target} & \textbf{\begin{tabular}[c]{@{}c@{}}Average\\ Accuracy\\ (DH)\end{tabular}} &
  \textbf{\begin{tabular}[c]{@{}c@{}}Mean\\ Average\\ Accuracy\end{tabular}} &
  \textbf{\begin{tabular}[c]{@{}c@{}}Median\\ Average\\ Accuracy\end{tabular}} &
  \textbf{\begin{tabular}[c]{@{}c@{}}Average\\ Accuracy\\ (no DH)\end{tabular}}\\
\hline \multirow{5}{*}{Word2Vec} & Fat        & 67.77 & 22.52 & 21.14 & 47.82\\
                          & Protein                     & 81.4  & 43.07 & 29.82 & 70.29\\
                          & Saturated fat & 85.85 & 15.92 & 27.36 & 75.71\\
                          & Sugars & 59.88 & 8.69  & 21.53 & 36.67\\
                          & Sodium         & 92.88 & 95.63 & 47.01 & 84.88\\
\hline \multirow{5}{*}{GloVe}    & Fat          & 67.14 & 22.52 & 21.14 & 47.04\\
                          & Protein                      & 81.07 & 43.07 & 29.82 & 69.99\\
                          & Saturated fat & 85.56 & 15.92 & 27.36 & 75.54\\
                          & Sugars  & 59.44 & 8.69  & 21.53 & 36.09\\
                          & Sodium   & 92.88 & 95.63 & 47.01 & 84.87\\
\hline \multirow{5}{*}{Doc2Vec}  & Fat  & 66.08 & 22.52 & 21.14 & 46.60\\
                          & Protein & 83.16 & 43.07 & 29.82 & 69.87\\
                          & Saturated fat & 85.02 & 15.92 & 27.36 & 75.24\\
                          & Sugars & 60.98 & 8.69  & 21.53 & 32.71\\
                          & Sodium & 92.72 & 95.63 & 47.01 & 84.88\\
\hline \multirow{5}{*}{BERT}     & Fat & 74.42 & 22.52 & 21.14 & 48.00\\
                          & Protein & 86.61 & 43.07 & 29.82 & 71.00\\
                          & Saturated fat & 87.69 & 15.92 & 27.36 & 75.44\\
                          & Sugars  & 75.86 & 8.69  & 21.53 & 35.91\\
                          & Sodium       & 92.78 & 95.63 & 47.01 & 84.87\\
\hline
\end{tabular}%
}
\end{table}

\begin{table}
\centering
\caption{Results for the Epicurious recipe dataset.}
\label{RepicuriousDH}
\resizebox{\columnwidth}{!}{
\begin{tabular}{|c|c|c|c|c|c|}
\hline
\textbf{Embedding} & \textbf{Target} & \textbf{\begin{tabular}[c]{@{}c@{}}Average\\ Accuracy\\ (DH)\end{tabular}} &
  \textbf{\begin{tabular}[c]{@{}c@{}}Mean\\ Average\\ Accuracy\end{tabular}} &
  \textbf{\begin{tabular}[c]{@{}c@{}}Median\\ Average\\ Accuracy\end{tabular}} &
  \textbf{\begin{tabular}[c]{@{}c@{}}Average\\ Accuracy\\ (no DH)\end{tabular}}\\
\hline \multirow{5}{*}{Word2Vec} & Fat & 54.04 & 19.22 & 12.89 & 33.06\\
                          & Protein  & 78.84 & 24.73 & 16.48 & 54.0\\
                          & Saturated fat & 86.05 & 18.92 & 16.03 & 75.96\\
                          & Sugars & 61.96 & 7.96  & 17.52 & 40.48\\
                          & Sodium & 90.52 & 43.66 & 44.99 & 82.34\\
\hline \multirow{5}{*}{GloVe}    & Fat  & 52.77 & 19.1  & 12.89 & 33.25\\
                          & Protein & 75.88 & 24.71 & 16.48 & 51.06\\
                          & Saturated fat & 86.03 & 18.9  & 16.04 & 76.04\\
                          & Sugars & 55.75 & 7.95  & 17.57 & 35.76\\
                          & Sodium  & 91.12 & 43.61 & 45.0 & 82.95 \\
\hline \multirow{5}{*}{Doc2Vec}  & Fat & 47.12 & 19.1  & 12.89 & 31.5\\
                          & Protein & 66.66 & 24.71 & 16.48 & 48.36\\
                          & Saturated fat & 84.34 & 18.9  & 16.04 & 75.66\\
                          & Sugars & 45.0 & 7.95  & 17.57 & 30.5\\
                          & Sodium & 90.56 & 43.61 & 45.0  & 82.84\\
\hline \multirow{5}{*}{BERT}     & Fat & 59.32 & 19.1 & 12.89 & 31.71\\
                          & Protein & 79.82 & 24.71 & 16.48 & 51.47 \\
                          & Saturated fat & 86.93 & 18.9  & 16.04 & 76.09\\
                          & Sugars & 60.85 & 7.95  & 17.57 & 37.56\\
                          & Sodium & 89.98 & 43.61 & 45.0 & 82.14\\

\hline
\end{tabular}%
}
\end{table}

\begin{table}
\centering
\caption{Results for the Salad recipes dataset.}
\label{RsaladrecipesDH}
\resizebox{\columnwidth}{!}{%
\begin{tabular}{|c|c|c|c|c|c|}
\hline
\textbf{Embedding} & \textbf{Target} & \textbf{\begin{tabular}[c]{@{}c@{}}Average\\ Accuracy\\ (DH)\end{tabular}} &
  \textbf{\begin{tabular}[c]{@{}c@{}}Mean\\ Average\\ Accuracy\end{tabular}} &
  \textbf{\begin{tabular}[c]{@{}c@{}}Median\\ Average\\ Accuracy\end{tabular}} &
  \textbf{\begin{tabular}[c]{@{}c@{}}Average\\ Accuracy\\ (no DH)\end{tabular}}\\
\hline \multirow{5}{*}{Word2Vec} & Fat & 44.98 & 17.25 & 12.33 & 29.11\\
                          & Protein & 69.79 & 21.91 & 16.0 & 49.84 \\
                          & Saturated fat  & 83.81 & 16.48 & 13.22 & 74.12 \\
                          & Sugars & 50.38 & 8.75  & 17.55 & 34.99\\
                          & Sodium & 88.06 & 22.39 & 44.85 & 76.01\\
\hline \multirow{5}{*}{GloVe}    & Fat & 44.02 & 17.25 & 12.33 & 28.44\\
                          & Protein & 66.96 & 21.91 & 16.00 & 46.39\\
                          & Saturated fat& 83.46 & 16.48 & 13.22 & 74.30\\
                          & Sugars & 45.67 & 8.75  & 17.55 & 32.49\\
                          & Sodium & 89.95 & 22.39 & 44.85 & 82.09\\
\hline \multirow{5}{*}{Doc2Vec}  & Fat & 40.56 & 17.25 & 12.33 & 26.21\\
                          & Protein & 59.23 & 21.91 & 16.0 & 41.66\\
                          & Saturated fat  & 82.44 & 16.48 & 13.22 & 73.5\\
                          & Sugars & 38.43 & 8.75  & 17.55 & 26.66\\
                          & Sodium & 90.76 & 22.39 & 44.85 & 82.46\\
\hline \multirow{5}{*}{BERT}     & Fat & 46.6  & 17.25 & 12.33 & 29.25\\
                          & Protein & 70.66 & 21.91 & 16.0 & 48.9\\
                          & Saturated fat  & 84.56 & 16.48 & 13.22 & 74.69\\
                          & Sugars & 51.86 & 8.75  & 17.55 & 33.2\\
                          & Sodium & 89.92 & 22.39 & 44.85 & 80.54\\

\hline
\end{tabular}%
}
\end{table}

\begin{table}
\centering
\caption{Results for the Yummly28K recipe dataset.}
\label{Ryummly28KDH}
\resizebox{\columnwidth}{!}{%
\begin{tabular}{|c|c|c|c|c|c|}
\hline
\textbf{Embedding} & \textbf{Target} & \textbf{\begin{tabular}[c]{@{}c@{}}Average\\ Accuracy\\ (DH)\end{tabular}} &
  \textbf{\begin{tabular}[c]{@{}c@{}}Mean\\ Average\\ Accuracy\end{tabular}} &
  \textbf{\begin{tabular}[c]{@{}c@{}}Median\\ Average\\ Accuracy\end{tabular}} &
  \textbf{\begin{tabular}[c]{@{}c@{}}Average\\ Accuracy\\ (no DH)\end{tabular}}\\
\hline \multirow{5}{*}{Word2Vec} & Fat            & 53.47 & 19.44 & 13.81 & 33.83\\
                          & Protein                      & 76.51 & 27.06 & 15.33 & 51.85\\
                          & Saturated fat  & 85.69 & 19.06 & 14.54 & 75.80 \\
                          & Sugars & 64.17 & 9.62  & 28.34 & 42.43\\
                          & Sodium          & 89.64 & 26.18 & 45.83 & 77.73 \\
\hline \multirow{5}{*}{GloVe}    & Fat             & 51.6  & 19.4  & 13.82 & 33.37\\
                          & Protein                      & 73.69 & 27.09 & 15.33 & 49.38\\
                          & Saturated fat  & 85.26 & 19.06 & 14.55 & 75.79\\
                          & Sugars  & 58.54 & 9.64  & 28.35 & 40.03\\
                          & Sodium           & 88.25 & 27.03 & 45.84 & 79.46 \\
\hline \multirow{5}{*}{Doc2Vec}  & Fat             & 44.79 & 19.4  & 13.82 & 31.71\\
                          & Protein                      & 60.52 & 27.09 & 15.33 & 45.84\\
                          & Saturated fat & 83.54 & 19.06 & 14.55 & 75.15\\
                          & Sugars  & 44.6  & 9.64  & 28.35 & 34.12 \\
                          & Sodium        & 87.26 & 27.03 & 45.84 & 79.91\\
\hline \multirow{5}{*}{BERT}     & Fat           & 57.36 & 19.4  & 13.82 & 33.20\\
                          & Protein                      & 77.26 & 27.09 & 15.33 & 48.26 \\
                          & Saturated fat & 86.58 & 19.06 & 14.55 & 75.80\\
                          & Sugars  & 65.31 & 9.64  & 28.35 & 38.82\\
                          & Sodium        & 82.92 & 27.03 & 45.84 & 75.98\\
\hline
\end{tabular}%
}
\end{table}

\begin{table}
\centering
\caption{Results for the Recipe box dataset.}
\label{RrecipeboxDH}
\resizebox{\columnwidth}{!}{%
\begin{tabular}{|c|c|c|c|c|c|}
\hline
\textbf{Embedding} & \textbf{Target} & \textbf{\begin{tabular}[c]{@{}c@{}}Average\\ Accuracy\\ (DH)\end{tabular}} &
  \textbf{\begin{tabular}[c]{@{}c@{}}Mean\\ Average\\ Accuracy\end{tabular}} &
  \textbf{\begin{tabular}[c]{@{}c@{}}Median\\ Average\\ Accuracy\end{tabular}} &
  \textbf{\begin{tabular}[c]{@{}c@{}}Average\\ Accuracy\\ (no DH)\end{tabular}}\\
\hline \multirow{5}{*}{Word2Vec} & Fat & 59.27 & 18.61 & 11.08 & 35.11 \\
                          & Protein & 80.68 & 26.53 & 18.82 & 58.32\\
                          & Saturated fat & 83.79 & 16.00 & 10.04 & 72.30\\
                          & Sugars & 59.02 & 9.61  & 12.82 & 35.48\\
                          & Sodium & 89.88 & 38.39 & 45.77 & 81.02\\
\hline \multirow{5}{*}{GloVe}    & Fat & 56.51 & 18.61 & 11.08 & 34.50\\
                          & Protein & 78.3  & 26.53 & 18.82 & 55.64\\
                          & Saturated fat  & 83.09 & 16.00  & 10.04 & 72.04 \\
                          & Sugars & 53.33 & 9.61  & 12.82 & 32.84 \\
                          & Sodium & 90.84 & 38.39 & 45.77 & 82.89\\
\hline \multirow{5}{*}{Doc2Vec}  & Fat & 51.17 & 18.61 & 11.08 & 32.51\\
                          & Protein & 70.94 & 26.53 & 18.82 & 51.94 \\
                          & Saturated fat  & 80.41 & 16.00  & 10.04 & 70.75\\
                          & Sugars & 42.91 & 9.61  & 12.82 & 27.24\\
                          & Sodium  & 90.13 & 38.39 & 45.77 & 82.22\\
\hline \multirow{5}{*}{BERT}     & Fat & 67.43 & 18.61 & 11.08 & 26.29\\
                          & Protein & 82.78 & 26.53 & 18.82 & 63.57\\
                          & Saturated fat  & 86.64 & 16.00  & 10.04 & 55.08\\
                          & Sugars  & 64.96 & 9.61  & 12.82 & 33.44\\
                          & Sodium  & 89.67 & 38.39 & 45.77 & 81.19\\
\hline
\end{tabular}%
}
\end{table}

\section*{Acknowledgment}
This research was supported by the Slovenian Research Agency (research core grant number P2-0098), and the European Union’s Horizon 2020 research and innovation programme (FNS-Cloud, Food Nutrition Security) (grant agreement 863059).
\bibliographystyle{IEEEtran}
\bibliography{bib}
\end{document}